\documentclass{article}

\usepackage{arxiv}
\usepackage{abstract}

\usepackage{babel}
\usepackage[utf8]{inputenc} 
\usepackage[T1]{fontenc}    
\usepackage{hyperref}       
\usepackage{url}            
\usepackage{booktabs}       
\usepackage{amsfonts}       
\usepackage{nicefrac}       
\usepackage{microtype}      
\usepackage{lipsum}
\usepackage{graphicx}
\graphicspath{ {./Images/} }

\usepackage{microtype}   
\usepackage{doi}
\usepackage{amsmath,amssymb,amsfonts,amsbsy}
\usepackage{algorithmic}
\usepackage{graphicx}
\usepackage{textcomp}
\usepackage{xcolor}
\usepackage{url}
\usepackage{multicol}
\usepackage{newtxtext,newtxmath}
\usepackage{soul}
\usepackage{float}
\usepackage{subfigure} 
\usepackage{adjustbox}
\usepackage{comment}
\usepackage{float}
\usepackage{siunitx}
\usepackage{cite}
\usepackage{multirow}
\usepackage{array} 
\usepackage{tikz}
\usetikzlibrary{shapes,arrows,shadows,positioning}
\usetikzlibrary{matrix,chains,positioning,decorations.pathreplacing}

\newcommand{\hlr}[1]{\textcolor{black}{#1}}

\title{Epanechnikov nonparametric kernel density estimation based feature-learning in respiratory disease chest X-ray images}

\author{
    Verónica Marsico, Antonio~Quintero-Rinc\'on\\
    Department of Data Science, Data Science and AI Laboratory.\\ Catholic University of Argentina (UCA), Argentina, Buenos Aires, Argentina \\
    \texttt{antonioquintero@uca.edu.ar}
\And
    Hadj~Batatia\\
    MACS School, Heriot-Watt University, Dubai Campus, United Arab Emirates.\\
\And
    To cite this work, please use this reference:\\
     Communications in Computer and Information Science (CCIS, Vol.2649, 2025))\\
    \doi{10.1007/978-3-032-06336-6\_3}}

\begin{document}
\maketitle
\begin{abstract}
This study presents a novel method for diagnosing respiratory diseases using image data. It combines ~Epanechnikov's ~non-parametric kernel density estimation (EKDE) with a bimodal logistic regression classifier in a statistical-model-based learning scheme. 
EKDE's flexibility in modeling data distributions without assuming specific shapes and its adaptability to pixel intensity variations make it valuable for extracting key features from medical images.
The method was tested on 13808 randomly selected chest X-rays from the COVID-19 Radiography Dataset, achieved an accuracy of 70.14\%, a sensitivity of 59.26\%, and a specificity of 74.18\%, demonstrating moderate performance in detecting respiratory disease while showing room for improvement in sensitivity.
While clinical expertise remains essential for further refining the model, this study highlights the potential of EKDE-based approaches to enhance diagnostic accuracy and reliability in medical imaging.

\keywords{Kernel density estimation (KDE) \and Epanechnikov \and Bimodal logistic regression.}
\end{abstract}

\section{Introduction}
\label{sec:intro}
Chest radiography is a widely available and economical diagnostic technique for imaging that uses a few radiation sources to create images of the internal structures of the chest. These images are invaluable for diagnosing and monitoring a wide range of conditions affecting the organs in the chest cavity, such as the heart and the lungs. This work focuses on respiratory diseases (RD) that can be diagnosed using a chest X-ray image. Most common diseases include asthma, chronic obstructive pulmonary disease, pneumonia, influenza, common colds, bronchitis, tuberculosis, and COVID-19. 
Early diagnosis of RD is imperative to provide appropriate treatment and prevent deaths.

Non-parametric kernel density estimators are widely used in the statistical analysis of sequences of random variables, providing a powerful tool when the data does not fit a known parametric distribution. The underlying idea of KDE (Kernel Density Estimation) is to estimate a probability density function (PDF) using another function called \hlr{k}ernel. 
The kernel function is scaled by an important bandwidth parameter $h$,  analogous to the histogram bin size. Classical kernels include the Gaussian, uniform, and triangle shapes. The Epanechnikov kernel has a parabolic shape, and a flexible structure, and stands out in modeling data that do not adhere to a particular pattern \cite{Lake2009}. In the medical setting, KDE has been used as a basis for a wide range of studies, highlighting its relevance and effectiveness in early disease detection, image processing, and clinical outcome analysis. To name a few examples, KDE has demonstrated its effectiveness and versatility in the early detection of Mild Cognitive Impairment (MCI), achieving an accuracy of $81.3\%$, and specificity of $82.7\%$ \cite{Veluppal2022}. Additionally, it has been successfully applied in the binarization of medical images \cite{Rodriguez2008}. Also, it was successfully applied in ultrasound imaging for hepatic steatosis, proving to be clinically valid, with a significant correlation between the KDE-based entropy parameter and liver fat measurements \cite{Gao2023}.

This work aims to determine if the Epanechnikov non-parametric kernel density estimation (EKDE) is a relevant feature extraction tool for a statistical-model-based learning method to detect abnormalities in chest X-ray images. The intention is not to compete with the forefront developments in CNN-based deep learning methods. The hypothesis to be tested is that EKDE can effectively model the pixel intensity distributions in X-ray images, allowing for an accurate representation of their features that can be used in a binary classification machine learning scheme. As far as we know, the Epanechnikov non-parametric kernel density estimation has not been investigated for the classification of chest X-ray images, which is the main contribution of this work.
In this study, the mean and standard deviation from the EKDE were calculated for 13808 randomly selected chest X-ray images from the publicly available COVID-19 Radiography Dataset, with $10192$ patients in normal conditions, and $3616$ were affected by COVID-19 (see Section \ref{ssec:db}). The resulting fitted distribution captured fine details in the images, particularly key features relevant to COVID-19. These two statistical features (mean and standard deviation) were then used as inputs to a bimodal logistic regression classifier, distinguishing normal images from those showing signs of disease. The method achieved an accuracy of 70.14\%, a sensitivity of 59.26\%, and a specificity of 74.18\%, indicating moderate effectiveness in identifying normal and diseased cases. These results highlight the utility of the proposed method for respiratory disease diagnosis in clinical settings. 

The rest of this document is organized as follows. Section \ref{sec:met} presents the proposed method in the following order: description of the data (Section \ref{ssec:db}), Epanechnikov non-parametric kernel density estimation (Section  \ref{ssec:kde}), feature vector parameters (Section \ref{ssec:feat}), classification and prediction model (Section \ref{ssec:reglog}) and performance metrics used. In Section \ref{sec:res}, results are analyzed and discussed. Finally, conclusions and perspectives are presented in Section. \ref{sec:con}.

\section{Methodology}
\label{sec:met}
\subsection{Database}
\label{ssec:db}
\begin{figure}[!hbt]
\centering
\subfigure[Patient in normal conditions]{\includegraphics[angle=0,width=1\textwidth]{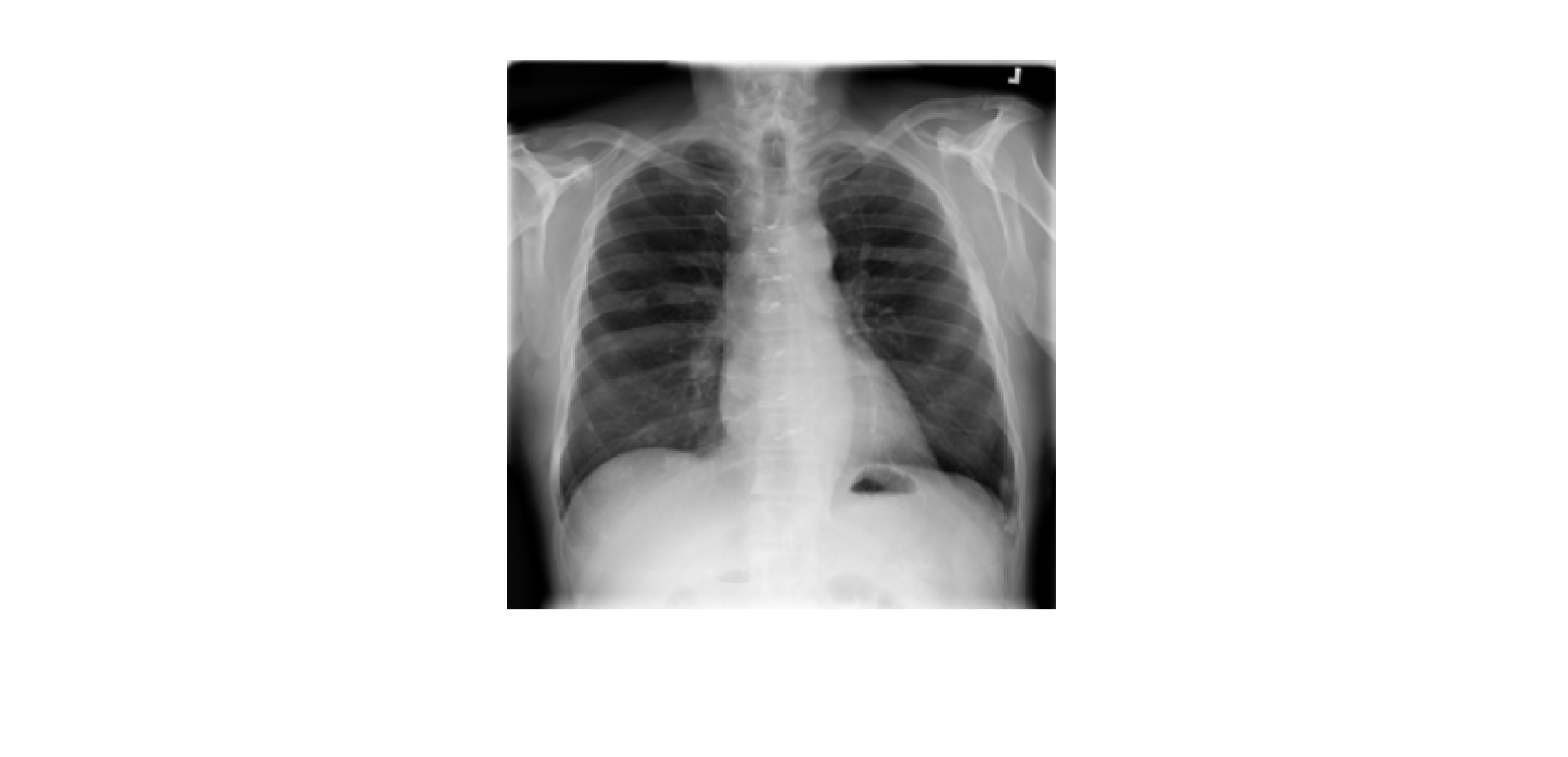}\label{fig:normal}}
\subfigure[Patient with COVID-19] {\includegraphics[angle=0,width=1\textwidth]{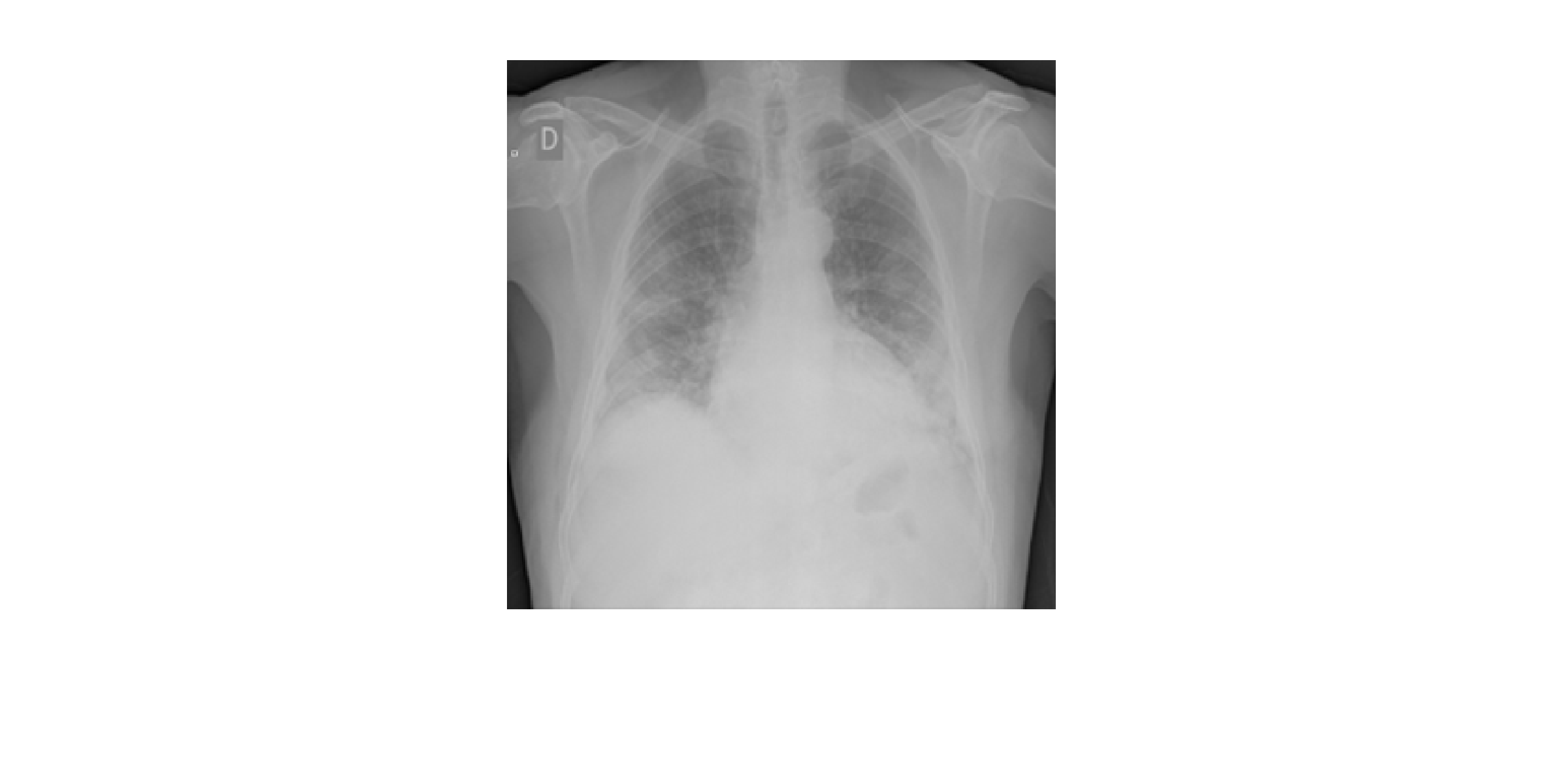} \label{fig:covid}}
\caption{Examples of chest X-ray (a)~{COVID-19}. (b)~{Normal}.}
\label{fig:images}
\end{figure}

The COVID-19 Radiography Database (DB) COVID-19 Chest X-ray images and Lung Masks Database, freely available in \cite{kaggle} was considered for experimentation. DB contains $10,192$ X-ray images corresponding to patients in normal conditions, $6,012$ images of pulmonary opacity (lung infection unrelated to COVID), $1,345$ images of viral pneumonia, and $3,616$ images of patients affected by COVID-19.  DB was compiled by Qatar University-Doha and the University of Dhaka-Bangladesh in collaboration with Pakistan and Malaysia.  See \cite{Chowdhury2020,Rahman2021} for more information about the complete dataset. 
It is important to clarify that the DB is challenging and complex. The DB contains several pneumonia cases that are not COVID-19, such as viral pneumonia, bronchitis, or healthy chest X-rays. The issue arises from the fact that the lung conditions of the COVID-19 patients in their image dataset can be particularly severe and pronounced, evident from the provided Figure~\ref{fig:images}. They can have conspicuous pneumonia patches, insufficient breath with numerous monitoring leads attached, diffuse or accentuated interstitial patterns, pulmonary opacities, interstitial infiltrates, or lack of aeration. \hlr{For more details, please consult \cite{QuinteroRincon2025a}}

\subsection{Epanechnikov non-parametric kernel density estimation (EKDE)}
\label{ssec:kde}
The non-parametric kernel density estimation for a random sample $x_1, x_2, \cdots, x_n$ drawn from a common and generally unknown density $f$ is given by: 

\begin{align}
\label{eq:kernel}
    \hat{f}(x)=\frac{1}{nh}\sum_{i=1}^n K\left ( \frac{x-x_i}{h}\right), ~~h>0
\end{align}
where $n$ is the sample size, $K$ is a kernel function, and $h$ is the bandwidth, also called the smoothing parameter or kernel bandwidth.\\
The parameters of the Epanechnikov kernel \cite{Epanechnikov1969} are defined as follows:
\begin{align}
\label{eq:kernelesp}
    K(x)  &=
   \left \{
      \begin{matrix} 
         \frac{3}{4}(1-x^2) & ~\text{ for } |x| < 1\\
         0           & ~\text{otherwise}
      \end{matrix}
   \right .\\
\label{eq:h}
   h &= \frac{0.9}{n^{\frac{1}{5}}}m, ~~m= \min\left( \sqrt{\sigma^2}, \frac{IQ}{1.349}\right)
\end{align}
where $\sigma^2$ is the variance and $IQ$ is the interquartile range. We refer the reader to \cite{Silverman1986, Gramacki2018} for a comprehensive treatment of the mathematical properties of kernel density estimation (KDE).

\subsection{Feature vector parameters}
\label{ssec:feat}

Let $X = \{x_1, x_2, \cdots, x_n\}$ be the \hlr{set of} chest X-ray images with COVID-19, corresponding to class $1$. Let $Y = \{y_1, y_2, \cdots, y_n$\} be the set of normal chest X-ray images, corresponding to class $0$.
Let $z \in X \cup Y$ be any image from these four sets, with $z \in \mathbb{R}^{N \times M}$, where $N$ is the number of rows and $M$ is the number of columns. Denoting the EKDE distribution of $z$ by $\hat{f}(z)$,  the feature vector of $z$ is: 
\begin{align}
    \label{eq:feat}
    \phi_{z} &= [\mu(\hat{f}(z)),\sigma(\hat{f}(z))] 
\end{align}
where $\mu$ and $\sigma$ are the mean and standard deviation, respectively. For simplicity, and without loss of generality, the feature vector for the image $x_i$ is denoted:
\begin{align}
    \label{eq:features}
    \phi_i =[\mu_i, \sigma_i]
\end{align}

\subsection{Bimodal logistic regression}
\label{ssec:reglog}

Consider a classification into two possible classes: $C_1$ for chest X-rays with COVID-19 and $C_2$ for normal chest X-rays. The posterior probability of class $C_1$ can be written as:

\begin{align}
\rho\left(C_1|x\right) &= \frac{\rho( x|C_1)\rho(C_1)}{\rho( x|C_1)p(C_1) + \rho( x|C_2)p(C_2)} 
\end{align}
By setting up 
\begin{align}
\alpha &= \ln \frac{\rho( x|C_1)\rho(C_1)}{\rho( x|C_2)\rho(C_2)},
\end{align}
we can easily show that the posterior reduces to the logistic function

\begin{align}
\rho\left(C_1|x\right) = S(\alpha)
= \frac{1}{1+\exp(-\alpha)}
\label{eq:sigmoid}
\end{align}
Consequently, assuming that all classes share the same covariance matrix $\Sigma$ \cite{Quintero2019}, we can write
\begin{align}
\label{eq:rhoC1}
\rho(C_1|\phi) &= S(\textnormal{w}^T \phi)\\
\label{eq:rhoC2}
\rho(C_2|\phi) &= 1 - \rho(C_1|\phi) \\
\textnormal{w} &= \Sigma^{-1}(\mu_1 - \mu_2)
\end{align}
 For a dataset $\left\{\left(\phi_i, t_i \right)\right\}$, where $t_i \in \{0,1\}$, with $C_1 = 0$ and $C_2 = 1$, the likelihood function can be written as:
\begin{align}
\label{eq:lhrl}
\rho(t | \textnormal{w}) = \prod_{i=1}^{n} y_i^{t_i} (1-y_i)^{1-t_i}
\end{align}
where $t = (t_1, t_2, \ldots, t_n)^T$ and $y_i = \rho(C_1|\phi_i)$.

\section{Results and Discussions}
\label{sec:res}

\begin{figure}[hbt!]
\centering
\subfigure[Normal ($\mu,\sigma$) EKDE parameters.] {\includegraphics[angle=0,width=60mm]{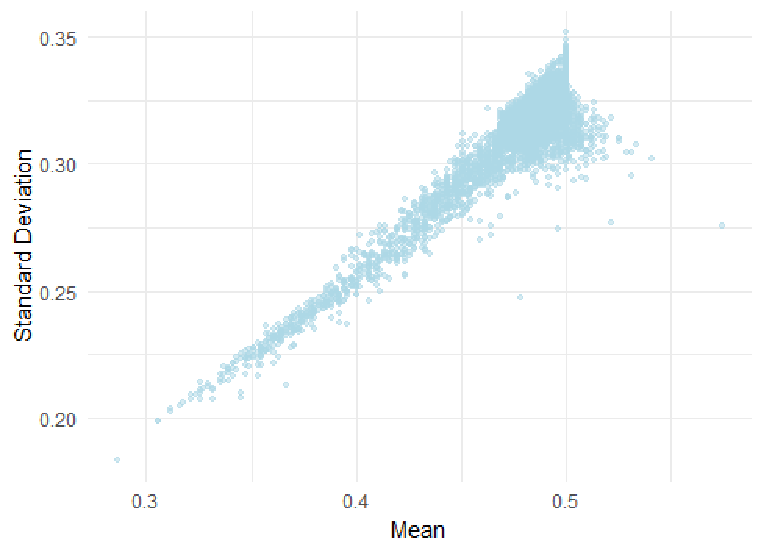}\label{fig:datan}}
\subfigure[COVID-19 ($\mu,\sigma$) EKDE parameters.]{\includegraphics[angle=0,width=60mm]{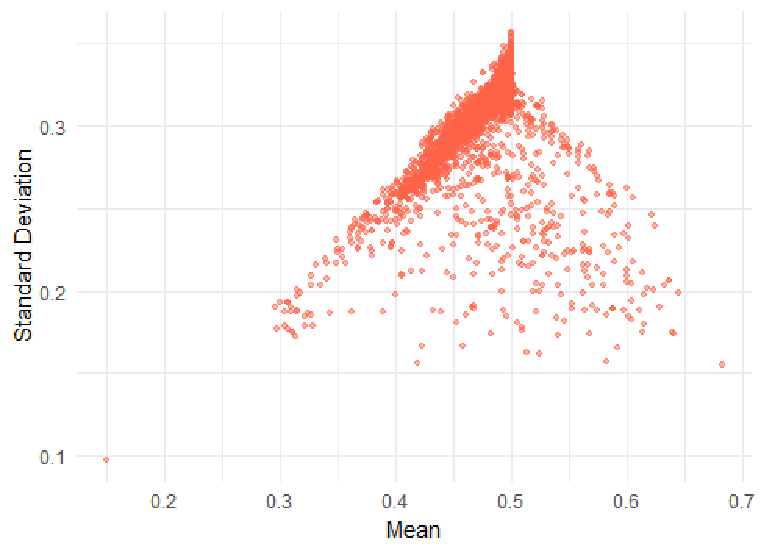}\label{fig:datac}}
\subfigure[Combined scatter plots of the two classes.] {\includegraphics[angle=0,width=75mm]{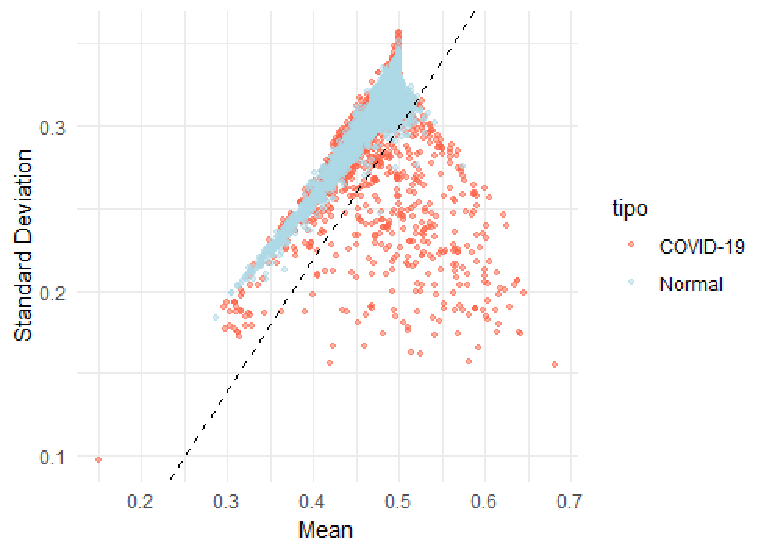}\label{fig:datacn}}
\caption{Scatter plots for the EKDE parameters ($\mu,\sigma$) for all classes. \ref{fig:datan} $10192$ normal CXR images, and \ref{fig:datac} $3616$ COVID-19 CXR images. \ref{fig:datacn}. Note that the COVID-19 images have a larger spread. The goal is to determine whether this separation is enough to discriminate between classes. \label{fig:datas}}
\end{figure}

\begin{figure}[hbt!]
\centering
\subfigure[] {\includegraphics[angle=0,width=0.61\textwidth]{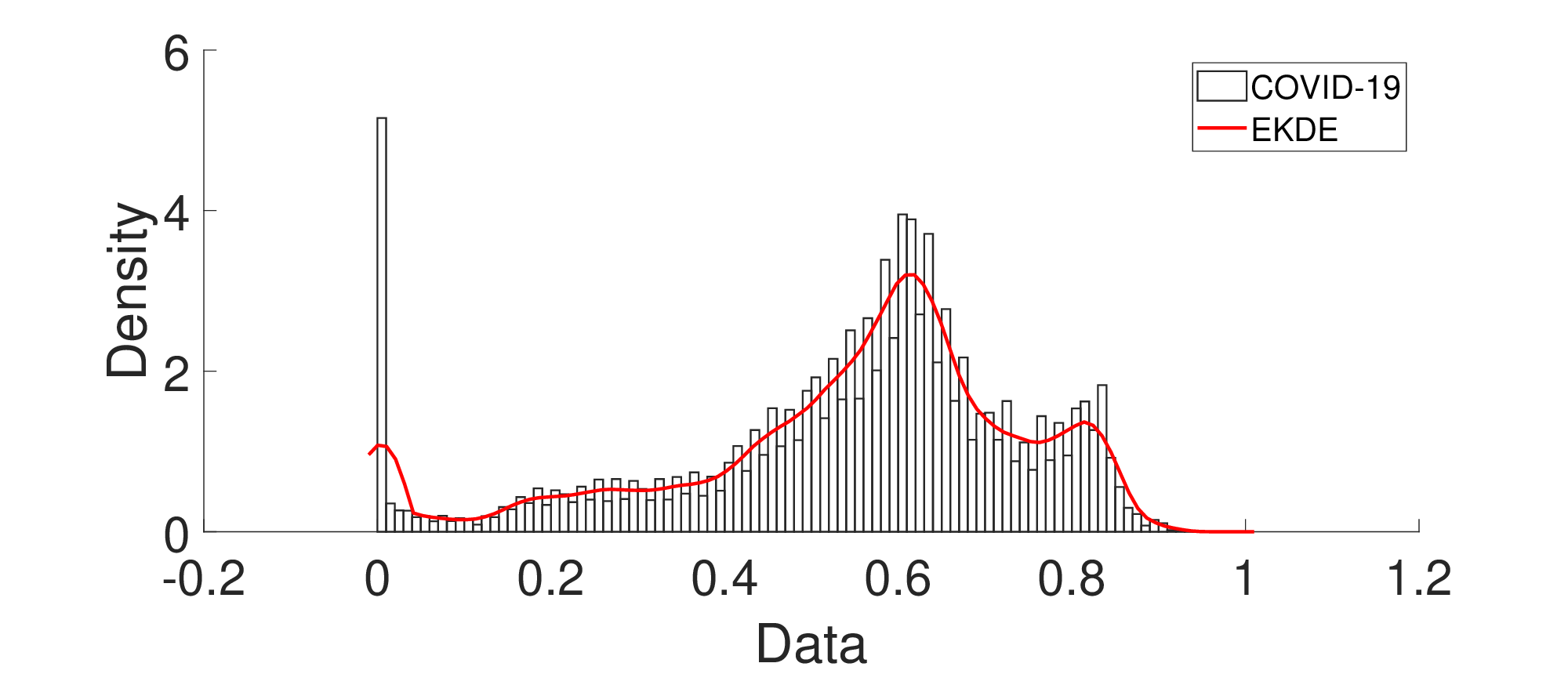} \label{fig:covidkernel1}}
\subfigure[] {\includegraphics[angle=0,width=0.61\textwidth]{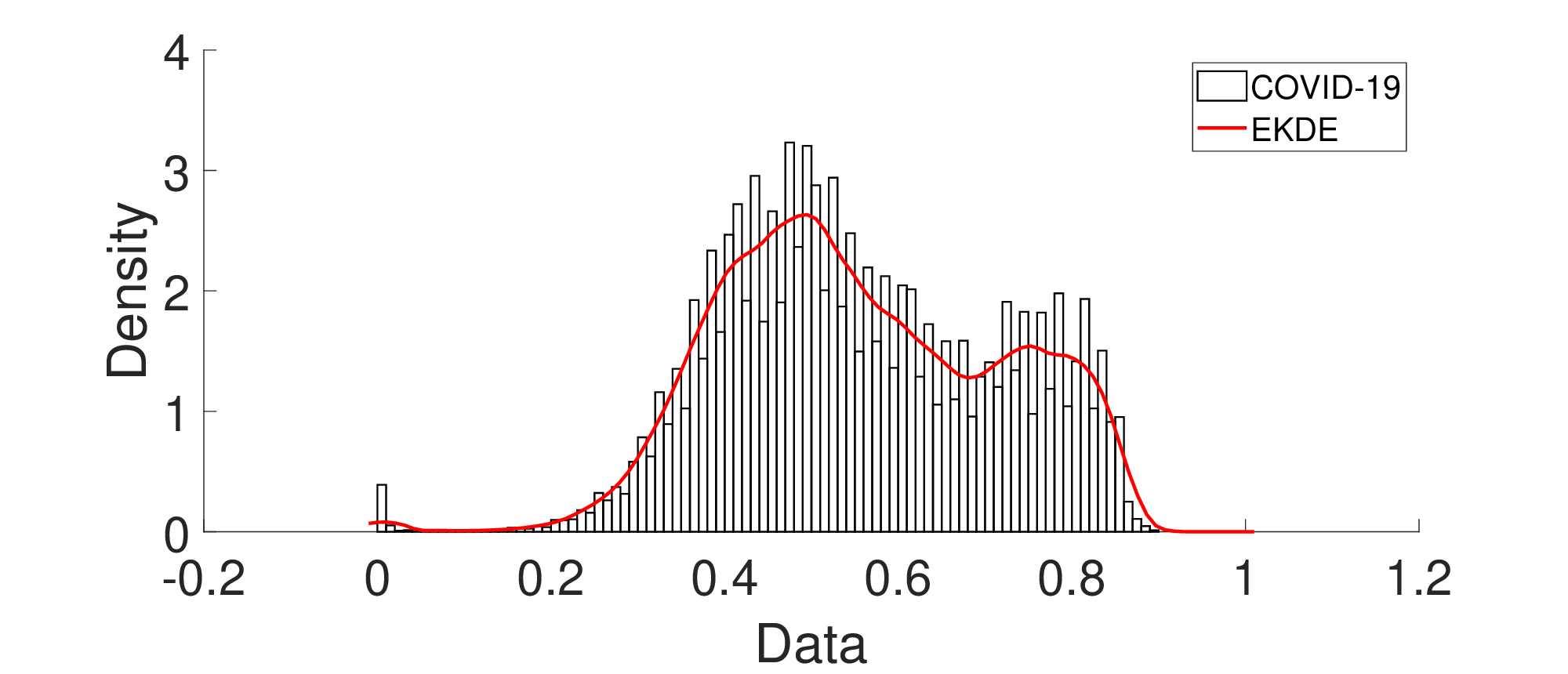} \label{fig:covidkernel2}}
\subfigure[] {\includegraphics[angle=0,width=0.61\textwidth]{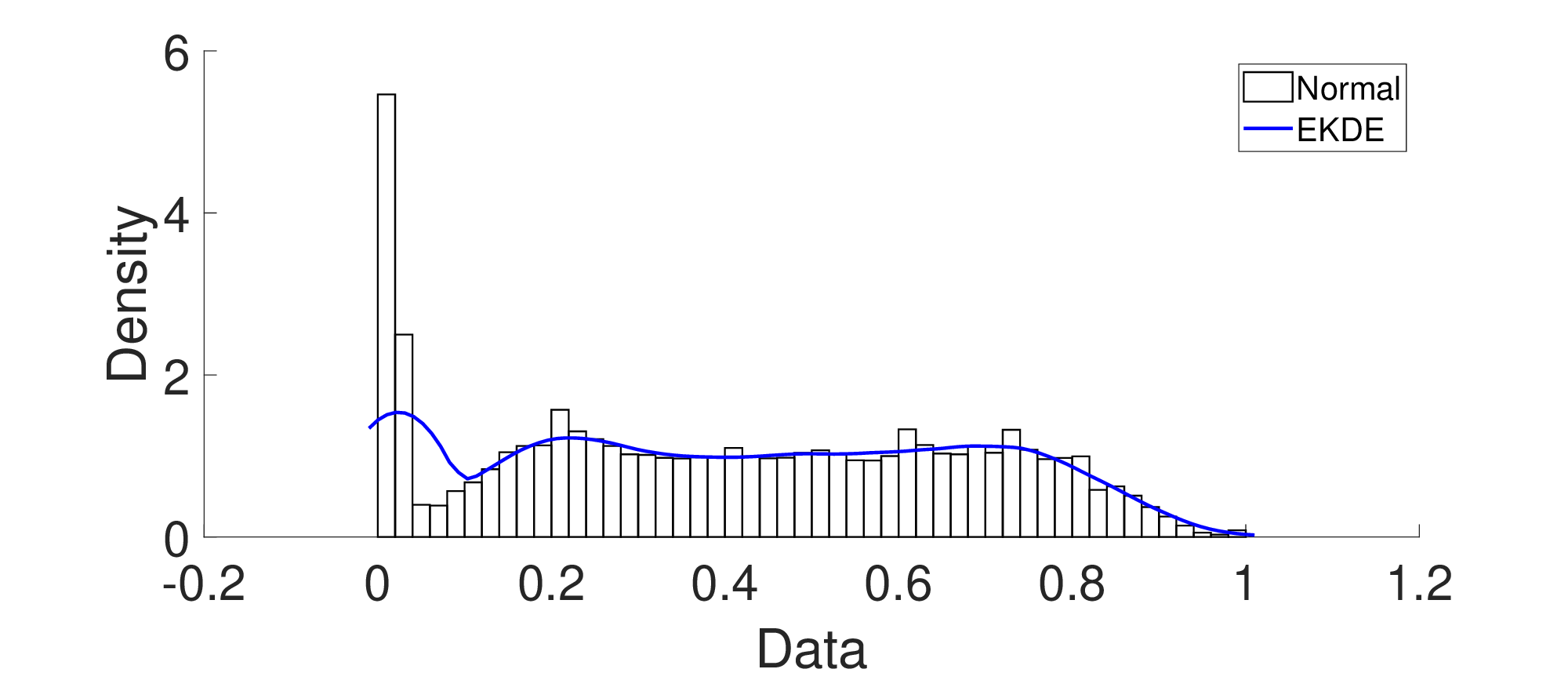} \label{fig:normalkernel1}}
\subfigure[] {\includegraphics[angle=0,width=0.61\textwidth]{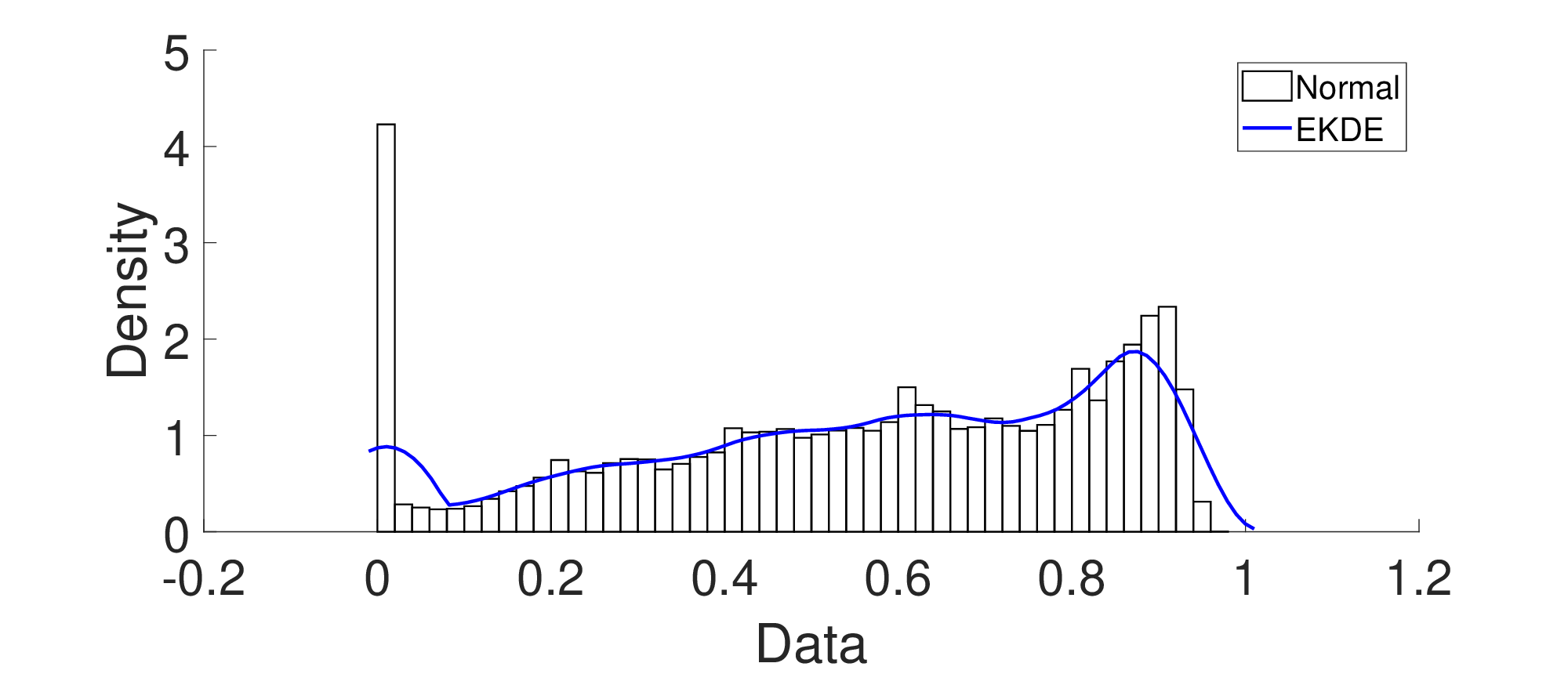} \label{fig:normalkernel2}}
\caption{EKDE fitting examples for the studied observations: (a-b)~COVID-19 CXR images, with its EKDE in red, (c-d)~Normal CXR images, with its EKDE in blue. Note that the histograms are bimodal but slightly different shapes, which we aim to capture using the classifier.}
\label{fig:kernels}
\end{figure}

It is well known that CNN-based deep learning methods are currently at the forefront in detecting anomalies in CXR images. The intention of this study is not to compete with these methods. This work aims to determine if the Epanechnikov nonparametric kernel density estimation (EKDE) is a relevant feature extraction tool for a statistical-model-based learning method to detect abnormalities in CXR images. For this purpose,  the mean  ($\mu$) and the standard deviation ($\sigma$) from EKDE were calculated as image features, yielding a feature vector given by equation \eqref{eq:features}, to be used in a bimodal logistic regression-based classifier, see section~\ref{ssec:reglog}. The results are reported below.\\
Figure~\ref{fig:datas} shows the scatter plots for the two EKDE parameters (mean, and standard deviation) observed through the complete CXR images dataset. All the images shared similar data points, therefore the superposition for all classes is evident. Figure~\ref{fig:datacn} shows the superposition of all scatter plots. Note that, it is difficult to discriminate between the two classes. The superposition observed for the mean between $0.3$ and $0.5$ and the standard deviation between $0.2$ and $0.35$ shows overlapping. However, the mean between $0.4$ and $0.5$ and the standard deviation between $0.2$ and $0.3$ split classes, through the binomial family, see equations \eqref{eq:rhoC1} and \eqref{eq:rhoC2}.\\
Figure~\ref{fig:kernels} shows examples of the Epanechnikov non-parametric kernel density estimate fitting for the observed CXR images. The distributions of the two classes are significantly different. The normal CXR image densities tend to be more uniform with a lower density amplitude than COVID-19 CXR images. The equation \eqref{eq:h} was used to calculate the minimum kernel bandwidth, resulting in a value of $h = 0.001$, Table~\ref{tab:h} shows the range and mean~$\pm$~standard deviation for the complete dataset. Note that the $h$ values differ only in their mean. This $h$ value plays a crucial role in the smoothness and adaptability of the resulting density curve. Specifically, it indicates a narrow bandwidth, yielding a more tailored estimation of data fluctuations. This means that $h$ may play an important role in capturing fine details such as pixel intensity in CXR images, as hypothesized particularly those related to the distinctive features of CXRs in the context of COVID-19 detection. Only a well-trained eye can distinguish these fluctuations for the CXR image examples, see Figure~\ref{fig:covid} and Figure~\ref{fig:normal}.

\begin{table}[hbt!]
\caption{kernel bandwidth $h$ for the two cases.}
\label{tab:h}
\centering
\begin{tabular}{|l|l|l|}
\hline
& \textbf{Bounds} & \textbf{mean$\pm$std} \\
\hline
COVID$-19$ & $[0.001, 0.108]$ & $0.024\pm0.008$ \\
\hline
normal    & $[0.001, 0.108]$ & $0.029\pm0.008$ \\
\hline
\end{tabular}
\end{table}
A preliminary analysis of all COVID-19 CXR images reveals a kernel mean of $0.48$, with a kernel standard deviation of $\pm 0.30$. In contrast, a kernel mean of $0.49$ with a standard deviation of $\pm 0.32$ is observed for all normal CXR images. 
Although these values are different and simultaneously very close, they suggest distinctive patterns in the density distribution in CXR images, highlighting the concentration around the kernel means and variability in the dispersion of these densities. This information is crucial for the characterization and thresholding of differentiating normal and disease-associated CXR images. The classes were formed under this criterion. Precisely, the feature vector of equation \eqref{eq:features} was used as the vector holding the two classes of CXR images, Class $1$ for COVID-19 and Class $0$ for normal, which feeds the bimodal logistic regression classifier model introduced in Section \ref{ssec:reglog}.

\begin{figure}[hbt!]
    \centering
    \subfigure[Training stage]{ \includegraphics[angle=0,width=0.7\textwidth]{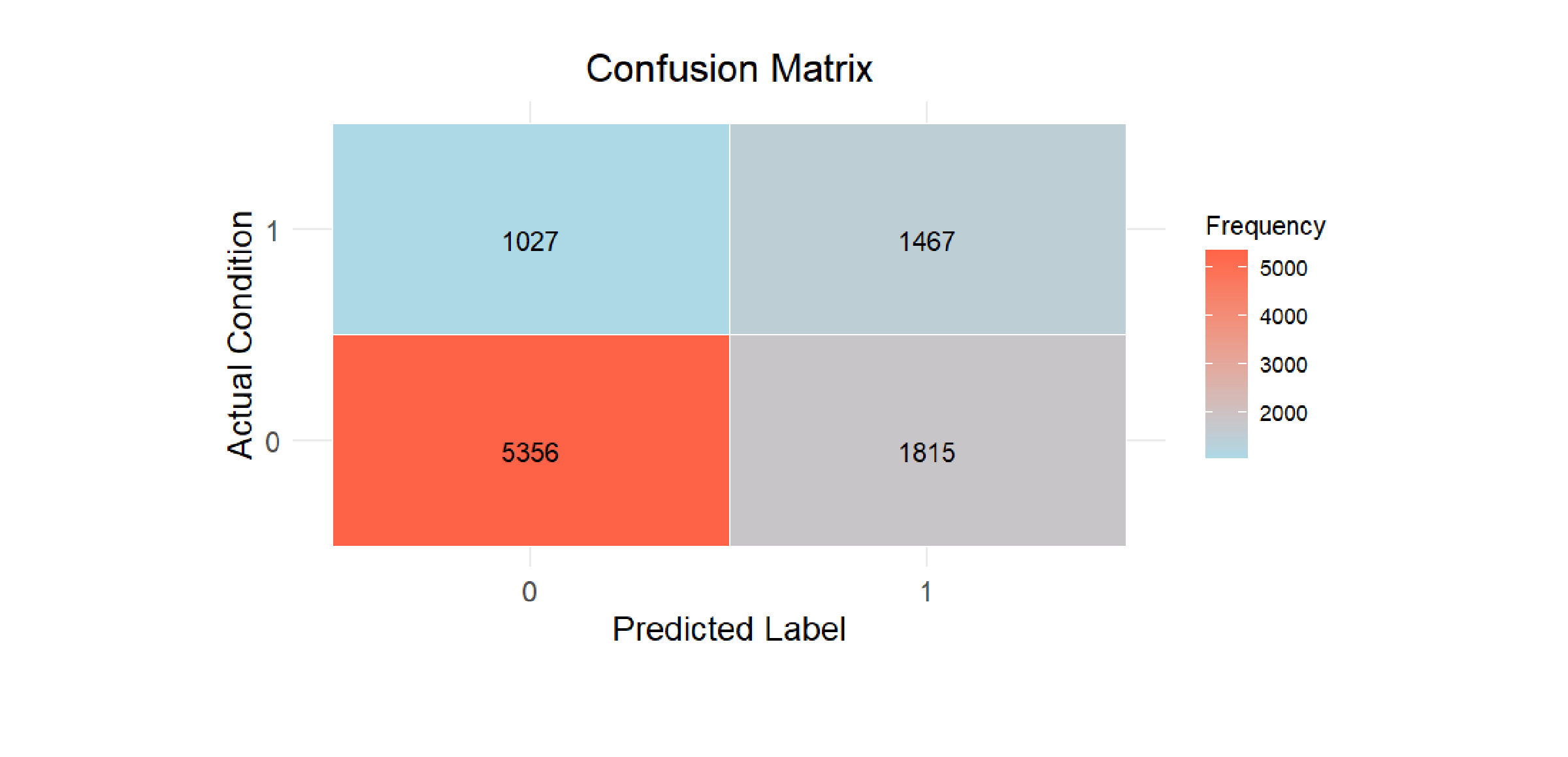}\label{fig:cmtrain}}
    \subfigure[Testing stage]{ \includegraphics[angle=0,width=0.7\textwidth]{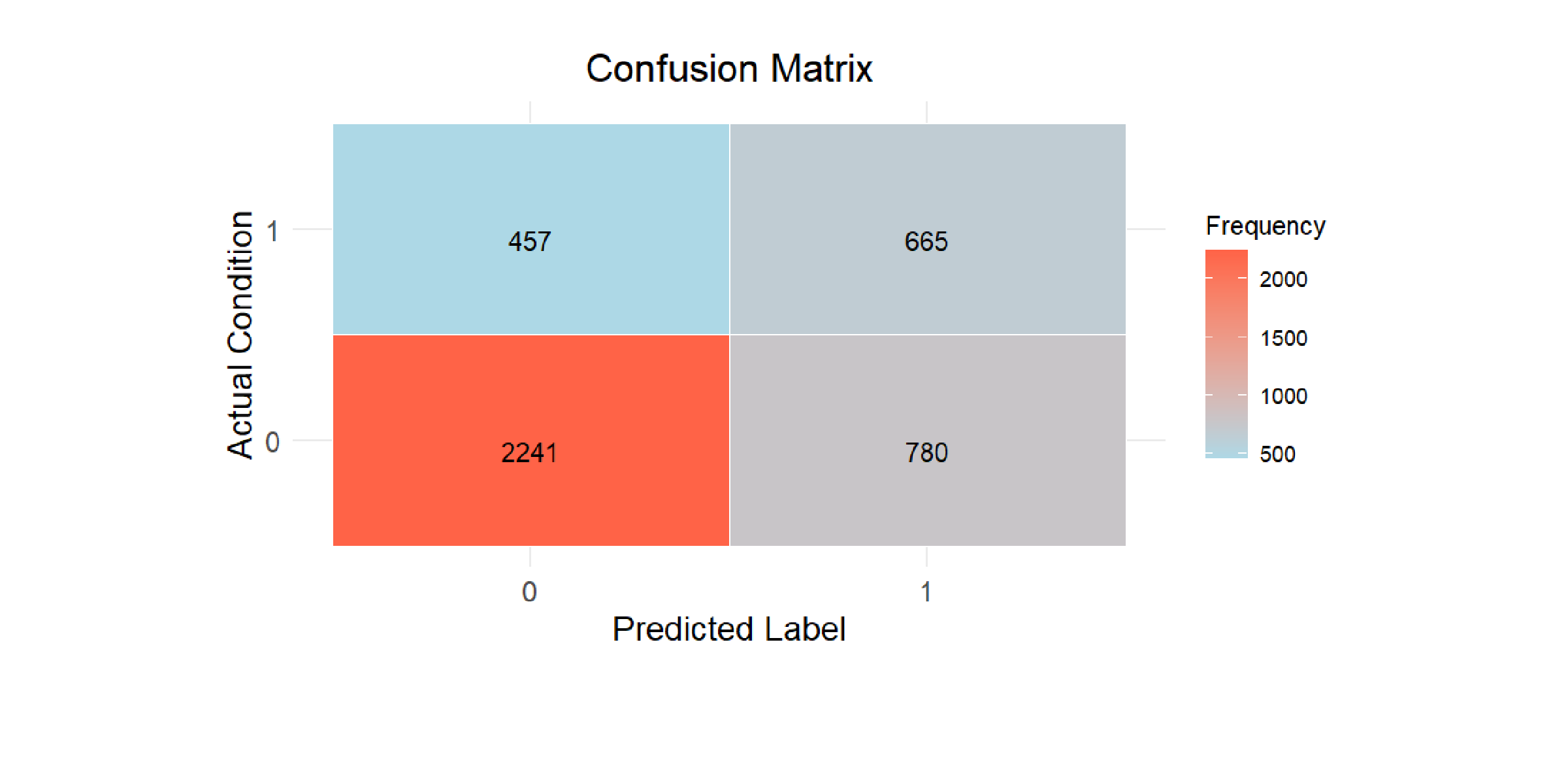}\label{fig:cmtest}}
    \caption{Confusion matrix of the proposed classifier in the training and testing stage. Class $1$ under $3616$ COVID-19 cases, and class $0$ under $10192$ normal cases.}
    \label{fig:mconf}
\end{figure}

For the bimodal logistic regression classifier model, $70\%$ of the dataset was used for the training stage, $7171$ for normal cases, and $2494$ for COVID-19 cases. The remaining $30\% $ were used in the testing stage, corresponding to $3021$ for normal cases, and $1122$ for COVID-19 cases. Both were random with a $ 10$-fold cross-validation, which was selected empirically. The model's effectiveness was evaluated using the ROC curve and the confusion matrix, as shown in Figure~\ref{fig:mconf}.
This classification model's area under the curve (AUC) is almost $0.7$ for the training and testing stages. Note that an AUC close to $1$ would first mean a greater model's ability to distinguish between patients with a positive COVID-19 diagnosis and those without the disease. Second, it would suggest an excellent balance between sensitivity and specificity.

Table~\ref{tab:confidence_intervals} shows the confidence intervals calculated to provide a certainty measure associated with the estimates of the bimodal logistic regression classifier model coefficients during the testing stage. These intervals reveal valuable information about the influence of specific features on the probability of belonging to the COVID-19 class. The intercept value ranges from $2.16$ to $3.58$, which suggests a radiological pattern associated with the pixel intensity in the CXR images. In this context, a high value indicates that the classifier model takes preference over detecting COVID-19 cases rather than missing them (false negatives).  Since minimizing undetected cases is crucial in detecting diseases such as COVID-19, this model behavior can be considered appropriate. Thus, a higher coefficient suggests that the classifier model tends to correctly classify COVID-19 cases, which is beneficial in a respiratory disease diagnosis.
\begin{table}[hbt!]
\caption{Confidence interval coefficients of the bimodal logistic regression classifier.}
\label{tab:confidence_intervals}
\centering
\begin{tabular}{|l|l|l|}
\hline
\textbf{Coefficient} & \textbf{$2.5\%$} & \textbf{$97.5\%$} \\
\hline
Intercept            & ~~$2.16$ & ~~$3.58$ \\
Mean                 & $20.63$  & $27.93$ \\
Standard Deviation    & -$54.76$ & -$44.99$ \\
\hline
\end{tabular}
\end{table}

\begin{figure}[hbt!]
    \centering
    \includegraphics[angle=0,width=0.55\textwidth]{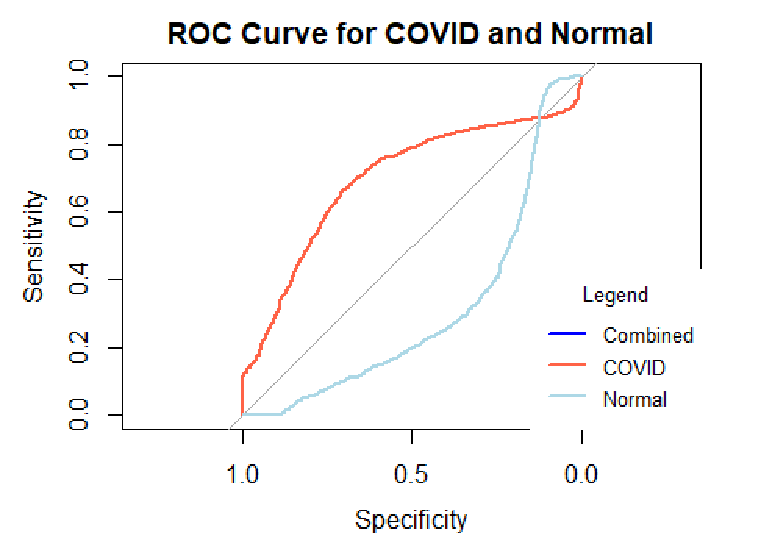}
    \caption{COVID-19 CXR images ROCs}
    \label{fig:rocs}
\end{figure}

\begin{figure}[hbt!]
    \centering
    \includegraphics[angle=0,width=0.75\textwidth]{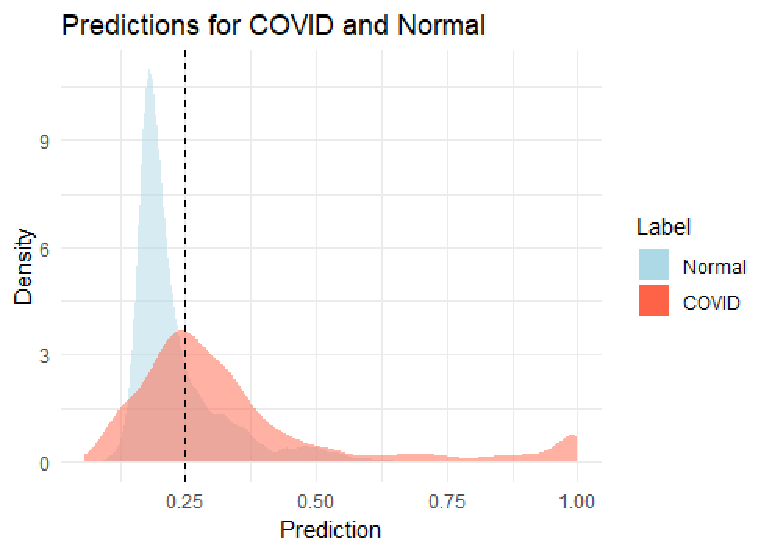}
    \caption{COVID-19 CXR images predictions}
    \label{fig:pred}
\end{figure}

Furthermore, the mean coefficient value changes from $20.63$ to $27.93$, suggesting that an increase in the mean density in CXR images is associated with a substantial increase in the log-odds of being classified as COVID-19. Similarly, the standard deviation coefficient value changes from -54.76 to -44.99, indicating an inverse relationship between the standard deviation and the log-odds of COVID-19. This implies that greater variability may reduce the probability of belonging to the COVID-19 class. These results provide key insights into how each variable contributes to the model predictions and are crucial for a detailed understanding of its performance.\\
The confidence bounds (95\%~CB) during the testing stage for the probabilities of each class were $[0.11, 0.99]$ for the COVID-19 class, and $[0.14, 0.50]$ for the normal class. These intervals play a crucial role in interpreting the predictions of a bimodal logistic regression classifier model in medical diagnosis settings. For the COVID-19 class, the interval suggests high confidence in COVID-19 predictions, indicating substantial accuracy in identifying positive cases. In contrast, the interval for the normal class reveals greater uncertainty and variability in predictions for this category. It is interesting to note that the normal class is within the bounds of the COVID-19 class, as evidenced by the scatter plots in Figure~\ref{fig:datas}. The COVID-19 class extends beyond the normal class bounds, which could be crucial for its detection in CXR images. This distinction underscores the model's robustness in identifying COVID-19 cases while highlighting the need for further attention or adjustment in classifying normal cases. These findings have significant implications for the clinical application of the model and emphasize the importance of considering the confidence associated with each prediction.\\
Figure~\ref{fig:mconf} shows the detailed confusion matrix of the proposed classifier for the training and testing stages. The following observations can be made in a medical setting: True Positives (TP) covering $1467$ for the training stage and $665$ for the testing stage, where the model has correctly identified patients with COVID-19, True Negatives (TN) covering  $5356$ for the training state and $2241$ for the testing stage, indicating the correct identification of normal individuals without COVID-19. The model exhibited robust performance in identifying positive cases. 
\hlr{The model demonstrated strong performance in identifying positive cases. However, False Negatives (FN) were found in $1027$ cases during the training stage and $457$ during the testing stage. These instances show where the model mistakenly predicted a negative outcome, even though the true outcome was positive}.
Finally, False Positives (FP) covering $1815$ for the training stage and $780$ for the testing stage, cases in which the model has incorrectly predicted the presence of COVID-19 in individuals who are not affected by the disease. The primary focus is reducing False Negatives (FN) ensuring effective detection of COVID-19 cases, given the clinical implications and the importance of preventing critical omissions in diagnosis \cite{Jacobi2020,Saurabha2022}. In addition, it may be the case where subsequent participation in disease is more likely among participants with false-positive results in an initial screening than patients with negative results \cite{Akira2016}. In contrast, the high specificity of $74.68\%$ for the training stage and $74.18\%$ for the testing stage indicates a strong ability to identify normal patients correctly. This finding reinforces the model's reliability in differential diagnosis situations \cite{MedicinaDiferencial2023}.\\
Likelihood ratios are essential in evaluating a classification model's effectiveness \cite{Walsh1969, Fischer2023}, particularly in the context of diagnostic tests, such as the one presented in this study. The positive likelihood ratio (LR+) with a value of $2.316$ indicates a reduced probability of obtaining a positive result in patients affected by the disease, highlighting the model's ability to accurately rule out positive cases. On the other hand, the negative likelihood ratio (LR-) with a value of $0.550$ suggests a substantially higher probability of obtaining a negative result in patients with COVID-19, emphasizing the model's ability to exclude those without the disease correctly. 
In particular, the low probability of obtaining a positive result (LR+) emphasizes the model's ability to rule out disease in patients without the condition, providing a valuable tool for accurate negative case detection.\\
These findings suggest that the proposed methodology provides a reasonable quantitative characterization of CXR images, allowing for moderate classification accuracy. Combining KDE analysis with the bimodal logistic regression classification model offers a promising tool for COVID-19 detection, subject to clinical validation. In the context of this study, an overall accuracy of $70.59\%$ for the training stage and $70.14\%$ for the testing stage of detecting COVID-19 cases is achieved. This result highlights the model's ability to make reasonable predictions on the test set. However, a more modest sensitivity of $58.88\%$ for the training stage and $59.26\%$ for the testing stage suggests that the model may miss many true cases of COVID-19. This consideration becomes critically important in the clinical setting, representing the major drawback of this method. Like any other artificial intelligence method,  the proposed method aids screening cases and must be accompanied by medical expertise \cite{Lamberti2023}. Overall, the model achieved for the training stage $70.59\%$ accuracy, $58.82\%$ True Positive Rate (sensitivity), $74.68\%$ True Negative Rate (specificity), $64.17\%$ F$_1$ score, $46.69\%$ Positive Predictive Values, and $83.91$ Negative Predictive Values; and for the testing stage $70.14\%$ accuracy, $59.26\%$ sensitivity, $74.18\%$ specificity, $64.24\%$ F$_1$ score, $46.02\%$ Positive Predictive Values, and $83.06\%$ Negative Predictive Values, see Figure~\ref{fig:rocs}.
In clinical settings, maximizing sensitivity is of fundamental importance to ensure the timely identification of COVID-19 cases, even at the cost of a higher number of false positives. This strategic approach underscores the priority of minimizing Type II errors (1-TPR), where the model fails to detect actual cases, compared to Type I errors (1-PPV), where the model detects non-existent cases. Additionally, the importance of addressing these ethical and practical considerations in the clinical deployment of the model is emphasized, focusing on maximizing sensitivity to prevent serious adverse outcomes.
To provide a clearer visualization of this situation, a density plot was generated with a threshold determined by the maximum density value for the COVID-19 class or label (see Figure~\ref{fig:pred}, note that by visual inspection, this maximum value is near $0.25$). This density plot illustrates the predictions generated by a bimodal logistic regression model in the context of detecting CXR images associated with COVID-19. Figure~\ref{fig:pred} was designed to offer a coherent visual representation: The $X$ axis is used to denote the model's predictions, ranging from values close to $1$, indicating high confidence in the positive class (COVID-19), to values close to $0$, reflecting greater confidence in the negative class (normal class).  The $Y$ axis represents the density of predictions at specific points on the $X$ axis, providing information on the frequency of occurrence of different prediction values. The colors in the plot distinguish between the actual classes, namely COVID-19 and Normal. A red dashed vertical line, located at the decision threshold of $0.25$, is incorporated to mark the classification boundary: to the left of this line, predictions are classified as normal, while to the right, as COVID-19. This graphical approach facilitates a detailed and quantitative assessment of the model's performance in detecting COVID-19 from CXR images.

Table~\ref{tab:comparative} shows some recent works focused on the COVID-19 Radiographic Database used in this work. It can be observed that the works are very diverse and dominated by deep learning techniques, where the detection of COVID-19 from CXR images is a field currently under research. This demonstrates that the present work is based on its wide use and relevance in the scientific community, besides its established utility in exploring diagnostic methods related to the respiratory disease under study.

\section{Conclusion}
\label{sec:con}

This work presented a new respiratory disease diagnosis method that employs the mean and the standard deviation from Epanechnikov's non-parametric kernel density estimation as features for a bimodal logistic regression classifier. This novel strategy was applied to $13808$ random CXR images from the well-known freely available COVID-19 Radiography Dataset achieving $70.14\%$ accuracy, $59.26\%$ sensitivity, and $74.18\%$ specificity. 

A detailed analysis of performance metrics reveals the presence of significant likelihood ratios: $2.31$ for the positive likelihood ratio ($LR+$), and $0.55$ for the negative likelihood ratio ($LR-$). These values support the model's ability to reduce the probability of obtaining a positive result in patients unaffected by the disease and increase the yielding likelihood of a negative result in patients with COVID-19, thus supporting the exclusion of negative cases.

\begin{table}[hbt!]
\caption{Comparison of some methods using the same database under study. CNN: Convolutional Neural Networks. $**$ The authors do not report any metric, only a statistical significance of $p < 0.00001$ in the lower lobes of the lungs.}
\label{tab:comparative}
\centering
\scalebox{0.75}{
\begin{tabular}{|p{2cm}|p{4cm}|p{4cm}|p{1cm}|p{7cm}|}
\hline
\textbf{Preprocessing} & \textbf{Methods} & \textbf{Features} & \textbf{ACC} & Ref. \\
\hline
- & EKDE + Bimodal logistic regression & Mean and standard deviation from EKDE & $70.14\%$ & This work\\
\hline
Resizing and augmentation of images & CNN & DenseNet & $99.70\%$ & \cite{Chowdhury2020}\\
\hline
- & CNN + SVM  & Mobilnetv2 & $98.50\%$ & \cite{Ullah2021}\\
\hline
- & CNN & DenseNet & 99.71\% & \cite{Kedia2021}\\
\hline
Gabor filtering & CNN & Xception & $94.13\%$ & \cite{Kumara2023} \\
\hline
- & Threshold-based binary segmentation to detect attenuation in the lobes & Proportion of white pixels
& ** & \cite{AlZyoud2023}\\
\hline
- & CNN & Inception V3, VGG16  & $98.00\%$ & \cite{Srinivas2023}  \\
\hline
Resizing of images & CNN & 32 filters of the convolutional layer & $97.43\%$ & \cite{Alablani2023}  \\
\hline
- & CNN & VGG16, VGG19& 92.00\% & \cite{Abdullah2024}\\
\hline
- & CNN & VGG19, EfficientNetB0 & 90.75\%& \cite{Houby2024}\\
\hline
- & CNN + SVM & VGG16, VGG19, MobileNet, Inception-v3, DenseNet201 & 92.00\% &\cite{Khoudour2024}\\
\hline
\end{tabular}}
\end{table}

The bimodal logistic regression classifier model suggests that the parameters of the distribution of the pixel intensity, estimated using Epanechnikov nonparametric kernel density estimation, can be used as features in a respiratory disease diagnosis method, as shown in the case of X-ray chest images for COVID-19.

The proposed method has three main advantages over the existing methods. First, this approach can be extended to work with other images, such as CT scans. Second, experimental methods can work without complex feature extraction techniques. Third, the computational complexity is low for univariate data and 2-D images $O(nm)$ compared with a CNN with a quadratic computational complexity $O(n^2)$.
The main limitation concerns the algorithms and practical solutions designed in the time-consuming context of KDE. They have a quadratic computational complexity $O(n^2)$, which can be an obstacle for large datasets, such as multivariate data. 

\hlr{Key areas can be explored in future work to improve and expand the application of the proposed methodology by integrating EKDE features as an interpretable branch in a CNN to test complementarity. This can be achieved using the early-fusion “dual-input” network technique, where the image pixels and the KDE feature vector are fed into separate branches of a CNN and then combined at a later stage \cite{Narayanan2020,Verma2025}. The identification of positive classes within imbalanced datasets using the metric of the area under the precision-recall curve (AUPRC) \cite{Beam2025}, particularly those involving rare disease cases, would be of considerable interest. AUPRC focuses exclusively on the performance of the positive class, thereby enhancing sensitivity to clinically significant advancements. It provides a more accurate portrayal of the model's effectiveness in detecting rare disease cases.} 
The optimization of the model's decision threshold also represents a promising area for future research \cite{Ramaswamy2025}. Adjusting this threshold could improve the model's sensitivity and specificity, allowing for a more precise adaptation to the target population's specific characteristics. Additionally, close collaboration with healthcare professionals to integrate clinical feedback into the continuous improvement process of the model would be essential for its successful implementation in real clinical settings.

Finally, continuously expanding and updating the training database could evaluate the model's adaptability to future virus variants. Together, these suggestions for future work can enhance the robustness and accuracy of the proposed model and contribute to the ongoing advancement in the detection and understanding of respiratory diseases, thus consolidating this study's scientific contribution to the field of diagnostic medicine.

\bibliographystyle{unsrt}

\begin{thebibliography}{10}

\bibitem{Lake2009}
Douglas~E. Lake.
\newblock Chapter 20 nonparametric entropy estimation using kernel densities.
\newblock In Michael~L. Johnson and Ludwig Brand, editors, {\em Computer Methods Part B}, volume 467 of {\em Methods in Enzymology}, pages 531--546. Academic Press, 2009.

\bibitem{Veluppal2022}
A.~Veluppal, D.~Sadhukhan, V.~Gopinath, and R.~Swaminathan.
\newblock Detection of mild cognitive impairment using kernel density estimation based texture analysis of the corpus callosum in brain {MR} images.
\newblock {\em IRBM}, 43(5):340--348, 2022.

\bibitem{Rodriguez2008}
Roberto Rodr\'iguez.
\newblock Binarization of medical images based on the recursive application of mean shift filtering: {Another} algorithm.
\newblock {\em Advances and Applications in Bioinformatics and Chemistry}, 1:1--12, 2008.

\bibitem{Gao2023}
Ruiyang Gao, Po-Hsiang Tsui, Shuicai Wu, Dar-In Tai, Guangyu Bin, and Zhuhuang Zhou.
\newblock Ultrasound entropy imaging based on the kernel density estimation: {A} new approach to hepatic steatosis characterization.
\newblock {\em Diagnostics}, 13(24), 2023.

\bibitem{kaggle}
{COVID-19} radiography database. {COVID-19} chest {X-ray} images and lung masks database.
\newblock \url{https://www.kaggle.com/tawsifurrahman/covid19-radiography-database?select={COVID-19}\_Radiography\_Dataset}.
\newblock Accessed: 2025-02-02.

\bibitem{Chowdhury2020}
Muhammad E.~H. Chowdhury, Tawsifur Rahman, Amith Khandakar, Rashid Mazhar, Muhammad~Abdul Kadir, Zaid~Bin Mahbub, Khandakar~Reajul Islam, Muhammad~Salman Khan, Atif Iqbal, Nasser~Al Emadi, Mamun Bin~Ibne Reaz, and Mohammad~Tariqul Islam.
\newblock Can {AI} help in screening viral and {COVID-19} pneumonia?
\newblock {\em IEEE Access}, 8, 2020.

\bibitem{Rahman2021}
Tawsifur Rahman, Amith Khandakar, Yazan Qiblawey, Anas Tahir, Serkan Kiranyaz, Saad~Bin {Abul Kashem}, Mohammad~Tariqul Islam, Somaya {Al Maadeed}, Susu~M. Zughaier, Muhammad~Salman Khan, and Muhammad~E.H. Chowdhury.
\newblock Exploring the effect of image enhancement techniques on {COVID-19} detection using chest {X-ray} images.
\newblock {\em Computers in Biology and Medicine}, 132:104319, 2021.

\bibitem{QuinteroRincon2025a}
Antonio Quintero-Rincón, Ricardo Di-Pasquale, Karina Quintero-Rodríguez, and Hadj Batatia.
\newblock Computer-based quantitative image texture analysis using multi-collinearity diagnosis in chest {X}-ray images.
\newblock {\em PLOS ONE}, 20(4):1--27, 04 2025.

\bibitem{Epanechnikov1969}
V.~A. Epanechnikov.
\newblock Non-parametric estimation of a multivariate probability density.
\newblock {\em Theory of Probability \& Its Applications}, 14(1):153--158, 1969.

\bibitem{Silverman1986}
B.W. Silverman.
\newblock {\em Density Estimation for Statistics and Data Analysis}.
\newblock Chapman and Hall, 1986.

\bibitem{Gramacki2018}
Artur Gramacki.
\newblock {\em Nonparametric Kernel Density Estimation and Its Computational Aspects}.
\newblock Springer, 2018.

\bibitem{Quintero2019}
Antonio Quintero~Rinc{\'o}n, M{\'a}ximo Flugelman, Jorge Prendes, and Carlos d'Giano.
\newblock Study on epileptic seizure detection in {EEG} signals using largest {Lyapunov} exponents and logistic regression.
\newblock {\em {Revista Argentina de Bioingenier{\'i}a}}, 23(2):17--24, 2019.

\bibitem{Jacobi2020}
Adam Jacobi, Michael Chung, Adam Bernheim, and Corey Eber.
\newblock Portable chest {X}-ray in coronavirus disease-19 ({COVID-19}): A pictorial review.
\newblock {\em Clinical Imaging}, 64:35--42, 2020.

\bibitem{Saurabha2022}
Nikitha Saurabha and Jyothi Shetty.
\newblock A review of intelligent medical imaging diagnosis for the {COVID-19} infection.
\newblock {\em Intelligent Decision Technologies}, 16(127):127--144, 2022.

\bibitem{Akira2016}
Akira Sato, Shota Hamada, Yuki Urashima, Shiro Tanaka, Hiroaki Okamoto, and Koji Kawakami.
\newblock The effect of false-positive results on subsequent participation in chest {X}-ray screening for lung cancer.
\newblock {\em Journal of Epidemiology}, 26(12):646--653, 2016.

\bibitem{MedicinaDiferencial2023}
F.~Javier~Laso Guzm\'an.
\newblock {\em Diagn\'ostico diferencial en medicina interna}.
\newblock Elsevier, 2023.

\bibitem{Walsh1969}
Thomas~J. Walsh, John~W. Garden, and Brian Gallagher.
\newblock Obliteration of retinal venous pulsations during elevation of cerebrospinal-fluid pressure.
\newblock {\em American Journal of Ophthalmology}, 67(6):954--956, 1969.

\bibitem{Fischer2023}
B.G. Fischer and A.T. Evans.
\newblock {SpPin} and {SnNout} are not enough. {I}t's time to fully embrace likelihood ratios and probabilistic reasoning to achieve diagnostic excellence.
\newblock {\em Journal of General Internal Medicine}, 38:2202--2204, 2023.

\bibitem{Lamberti2023}
William~Franz Lamberti.
\newblock An overview of explainable and interpretable {AI}.
\newblock In Feras~A. Batarseh and Laura~J. Freeman, editors, {\em {AI} Assurance}, pages 55--123. Academic Press, 2023.

\bibitem{Ullah2021}
Naeem Ullah and Ali Javed.
\newblock Deep features comparative analysis for {COVID-19} detection from the chest radiograph images.
\newblock In {\em 2021 International Conference on Frontiers of Information Technology (FIT)}, pages 258--263, 2021.

\bibitem{Kedia2021}
Priyansh Kedia, Anjum, and Rahul Katarya.
\newblock Covnet-19: A deep learning model for the detection and analysis of {COVID-19} patients.
\newblock {\em Applied Soft Computing}, 104:107184, 2021.

\bibitem{Kumara2023}
Chamoda~Tharindu Kumara, Sandunika~Charuni Pushpakumari, Ashmini~Jeewa Udhyani, Mohamed Aashiq, Hirshan Rajendran, and Chinthaka~Wasantha Kumara.
\newblock Image enhancement {CNN} approach to {COVID-19} detection using chest {X-ray} images.
\newblock {\em Engineering Proceedings}, 55(1), 2023.

\bibitem{AlZyoud2023}
W.~Al-Zyoud, D.~Erekat, and R.~Saraiji.
\newblock {COVID-19} chest {X-ray} image analysis by threshold-based segmentation.
\newblock {\em Heliyon}, 9(3):e14453, 2023.

\bibitem{Srinivas2023}
K.~Srinivas, R.~Gagana~Sri, K.~Pravallika, K.~Nishitha, and S.R. Polamuri.
\newblock {COVID-19} prediction based on hybrid inception {V3} with {VGG16} using chest {X-ray} images.
\newblock {\em Multimedia Tools and Applications}, 2023.

\bibitem{Alablani2023}
Ibtihal A.~L. Alablani and Mohammed J.~F. Alenazi.
\newblock {COVID-ConvNet}: A convolutional neural network classifier for diagnosing {COVID-19} infection.
\newblock {\em Diagnostics}, 13(10), 2023.

\bibitem{Abdullah2024}
Mohan Abdullah, Ftsum berhe Abrha, Beshir Kedir, and Takore~Tamirat Tagesse.
\newblock A hybrid deep learning {CNN} model for {COVID-19} detection from chest {X}-rays.
\newblock {\em Heliyon}, 10(5):1--13, 2024.

\bibitem{Houby2024}
Enas M. F.~El Houby.
\newblock {COVID‑19} detection from chest {X}-ray images using transfer learning.
\newblock {\em Scientific Reports}, 14(1):1--13, 2024.

\bibitem{Khoudour2024}
Mohamed~Elamine Khoudour, Isma{\"i}l Biskri, and Nadia Ghazzali.
\newblock {COVID-19} detection based on deep features and {SVM}.
\newblock In {\em Computational Collective Intelligence}, pages 107--119, 2024.

\bibitem{Narayanan2020}
Athma Narayanan, Avinash Siravuru, and Behzad Dariush.
\newblock Gated recurrent fusion to learn driving behavior from temporal multimodal data.
\newblock {\em IEEE Robotics and Automation Letters}, 5(2):1287--1294, 2020.

\bibitem{Verma2025}
Akash Verma and Arun~Kumar Yadav.
\newblock {FusionNet: D}ual input feature fusion network with ensemble based filter feature selection for enhanced brain tumor classification.
\newblock {\em Brain Research}, 1852:149507, 2025.

\bibitem{Beam2025}
Colin Beam.
\newblock Resolving power: a general approach to compare the distinguishing ability of threshold-free evaluation metrics.
\newblock {\em Machine Learning}, 114(9), 2025.

\bibitem{Ramaswamy2025}
Ramesh~Kumar Ramaswamy, Pannangi Naresh, Chilamakuru Nagesh, and Santhosh~Kumar Balan.
\newblock Multilevel thresholding technique with {A}rchery {G}old {R}ush {O}ptimization and{ PCNN}-based childhood medulloblastoma classification using microscopic images.
\newblock {\em Biomedical Signal Processing and Control}, 107:107801, 2025.

\end{thebibliography}

\end{document}